\pdfoutput=1
\documentclass[11pt,a4paper]{article}
\usepackage{array}
\usepackage[footnotesize,bf]{caption}
\usepackage[dvips]{graphicx}
\usepackage{color} 
\usepackage{amsmath}
\usepackage{fancyhdr}
\usepackage[]{natbib}
\usepackage[LGRgreek]{mathastext}
\usepackage[yyyymmdd,hhmmss]{datetime}

\fancyheadoffset[R]{0pt}
\fancyfootoffset[R]{0pt}

\definecolor{orange}{rgb}{1.0,0.5,0.0}
\definecolor{aqgr}  {rgb}{0.0,1.0,0.6} 
\definecolor{viol}  {rgb}{0.8,0.6,0.8}
\definecolor{figdr} {rgb}{1.0,1.0,1.0} 
\definecolor{colnu} {rgb}{1.0,0.0,1.0} 
\definecolor{colhd} {rgb}{1.0,0.8,0.0} 

\newcolumntype{C}[1]{>{\centering\let\newline\\\arraybackslash\hspace{0pt}}m{#1}}
\newif\ifhpar

\title{\vspace{-0.5cm}\bfseries{\textsc{An artificial neural network to find \\
   correlation patterns in \\ an arbitrary number of variables}}}

\author{Alessandro Fontana}
\date{}

\begin{document}
\maketitle
   
\clubpenalty=10000
\widowpenalty=10000

\begin{abstract}
Methods to find correlation between variables are of interest to many disciplines, including statistics, machine learning, (big) data mining and neurosciences. Parameters that measure correlation between two variables are of limited utility when used with multiple variables. In this work, I propose a simple criterion to measure correlation among an arbitrary number of variables, based on a dataset. The central idea is to i) design a function of the variables that can take different forms depending on a set of parameters, ii) calculate the difference between a statistic associated to the function computed on the dataset and the same statistic computed on a randomised version of the dataset, called ``scrambled'' dataset, and iii) optimise the parameters to maximise this difference. Many such functions can be organised in layers, which can in turn be stacked one on top of the other, forming a neural network. The function parameters are searched with an enhanced genetic algorithm called POET and the resulting method is tested on a cancer gene dataset. The method may have potential implications for some issues that affect the field of neural networks, such as overfitting, the need to process huge amounts of data for training and the presence of ``adversarial examples''.  
\end{abstract}


\section{Correlation measures}  

\ifhpar \colorbox{colhd}{introduction} \\ \fi
This paper is concerned with a new method to discover correlation patterns in an arbitrary number of variables, based on a dataset. The importance of the problem cannot be overstated: it can be argued that finding correlations in a set of variables is indeed the central objective of intelligence. The problem is relevant for many fields, such as statistics, machine learning, data mining and neurosciences. The method that we are going to propose lies at the intersection of all these disciplines.

\ifhpar \colorbox{colhd}{correlation and MI} \\ \fi
Many methods exist to measure correlation between variables. Most of these measures are based on some kind of comparison between the joint probability of the variables and the product of their marginal probabilities. Two such measures are represented by Pearson correlation \citep{pearson1895} and by mutual information \citep{cover1991} defined in formulas (1) and (2) (the correlation formula refers to the case of binary variables). 

\begin{table}[h]
\vskip 0.25cm
\center{
\begin{tabular}{C{8cm} C{8cm}}
$cor(A,B)= \frac {P(A,B) - P(A)P(B)} {\sqrt{P(A)(1-P(A))P(B)(1-P(B))}} $ (1) &
$I(A,B)= \displaystyle\sum_{a \in A} \displaystyle\sum_{b \in B} P(a,b)$ 
$log \left( \frac{P(a,b)}{P(a)P(b)} \right) $ (2)
\end{tabular}}
\vskip 0.25cm
\end{table}

\ifhpar \colorbox{colhd}{covariance matrix} \\ \fi
These measures have an intrinsic limit: they are able to capture correlation between only two variables. The covariance matrix extends the concept to an arbitrary number of variables, by providing the covariances between all possible couples of variables (matrix element (i,j) represents the covariance between variables i and j): however, this is not a measure of correlation among all variables \emph{considered together}. 

\ifhpar \colorbox{colhd}{open problem} \\ \fi
An example can help to clarify the issue. Let us imagine to have a dataset with 7 binary variables and let us assume the existence of the following ``rule'': if the first six variables are TRUE, then the 7th variable is TRUE, while if any of the first six variables is FALSE, then the 7th variable is FALSE. This ``perfect'' correlation is very unlikely to be captured through a pairwise correlation analysis.

\ifhpar \colorbox{colhd}{no arbitrary function} \\ \fi
There would be a perfect correlation between the 7th variable and another variable defined as the logical \textit{And} of the first six, but this variable does not exist. Formulas (1) and (2) are based on a comparison between the product of marginal probabilities and the probability of $A \cap B $. This is indeed a function of variables A and B, but it is just a special case: why not comparing $P(A) \cdot P(B)$ to P($A \cup B$), or any other function?  We need to define arbitrary functions of the variables and look for those with the highest correlation value. We need a generalised definition of correlation.

\ifhpar \colorbox{colhd}{paper structure} \\ \fi
The paper is organised as follows: section 2 introduces a new criterion to measure correlation; section 3 describes how to build a neural network based on this criterion and defines an objective functions for each network layer; section 4 describes the method used to train the function parameters and the simulation carried out to test the method; section 5 discusses some possible implications for other disciplines; section 6 draws the conclusions and outline future research directions.

\section{Hyper-occurrence}  

\begin{figure}[t]
\begin{minipage}[t]{0.47\textwidth} \centering \hspace*{-0.0cm}
\includegraphics[width=0.95\textwidth]{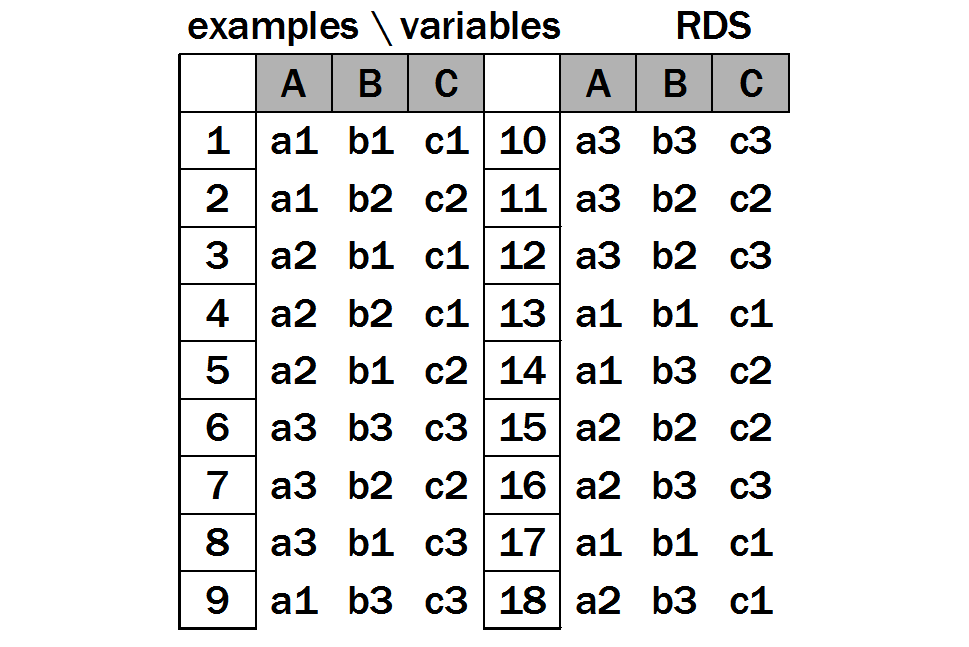}
\caption{Real dataset (RDS). Example of dataset with three variables (A,B,C) that can take certain values (\{a1,a2,a3\}, \{b1,b2,b3\}, \{c1,c2,c3\} respectively), each with equal probability. Some value combinations, e.g. (a1,b1,c1), ``hyper-occur''.}
\label{rdsdef}
\end{minipage} \qquad
\begin{minipage}[t]{0.47\textwidth} \centering \hspace*{-0.0cm}
\includegraphics[width=0.95\textwidth]{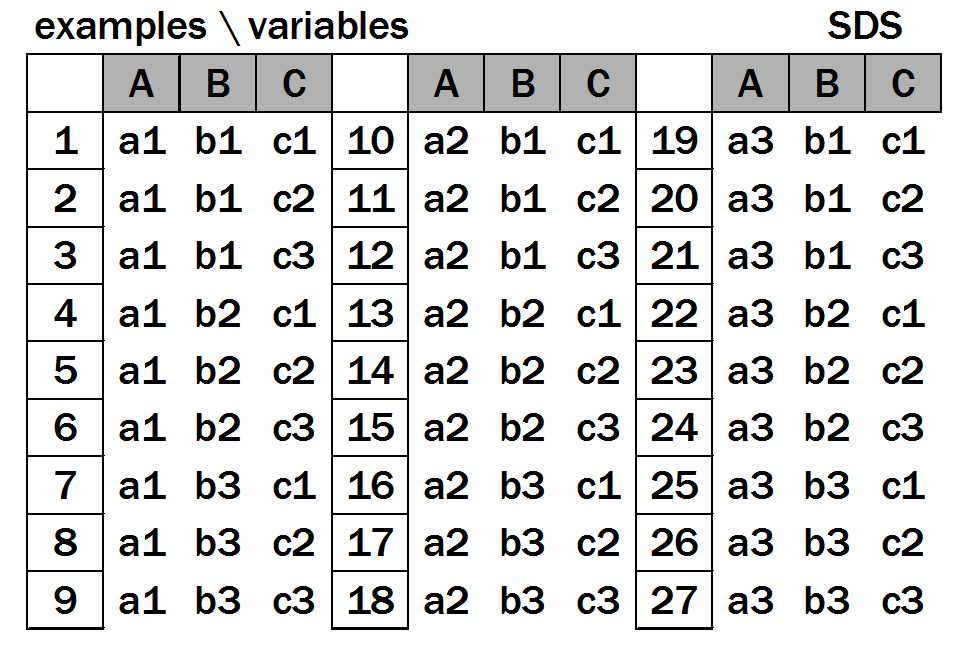}
\caption{Scrambled dataset (SDS). The scrambled dataset is a randomised version of the real dataset obtained combining the possible values of each variable in all possible ways.}
\label{sdsdef}
\end{minipage} \qquad
\end{figure}

\ifhpar \colorbox{colhd}{RDS definition} \\ \fi
Let us reformulate the correlation problem in more general terms. Let us suppose to have a dataset composed of a number of examples on a set of variables which can take real values in the [0,1] interval and have arbitrary probability distributions. Based on the data, we want to find correlation patterns in the variables. 

\ifhpar \colorbox{colhd}{central idea} \\ \fi
The idea to generalise correlation is to i) design a function of the dataset variables that can take different forms depending on a set of parameters, ii) calculate a score based on a comparison between a statistic associated to the function computed on the ``real'' dataset (RDS) and the same statistic computed on a randomised version of the dataset, called ``scrambled'' dataset (SDS) and iii) optimise the parameters to maximise the score. 

\begin{figure}[t] \begin{center}
{\fboxrule=0.0mm\fboxsep=0mm\fbox{\includegraphics[width=11.00cm]{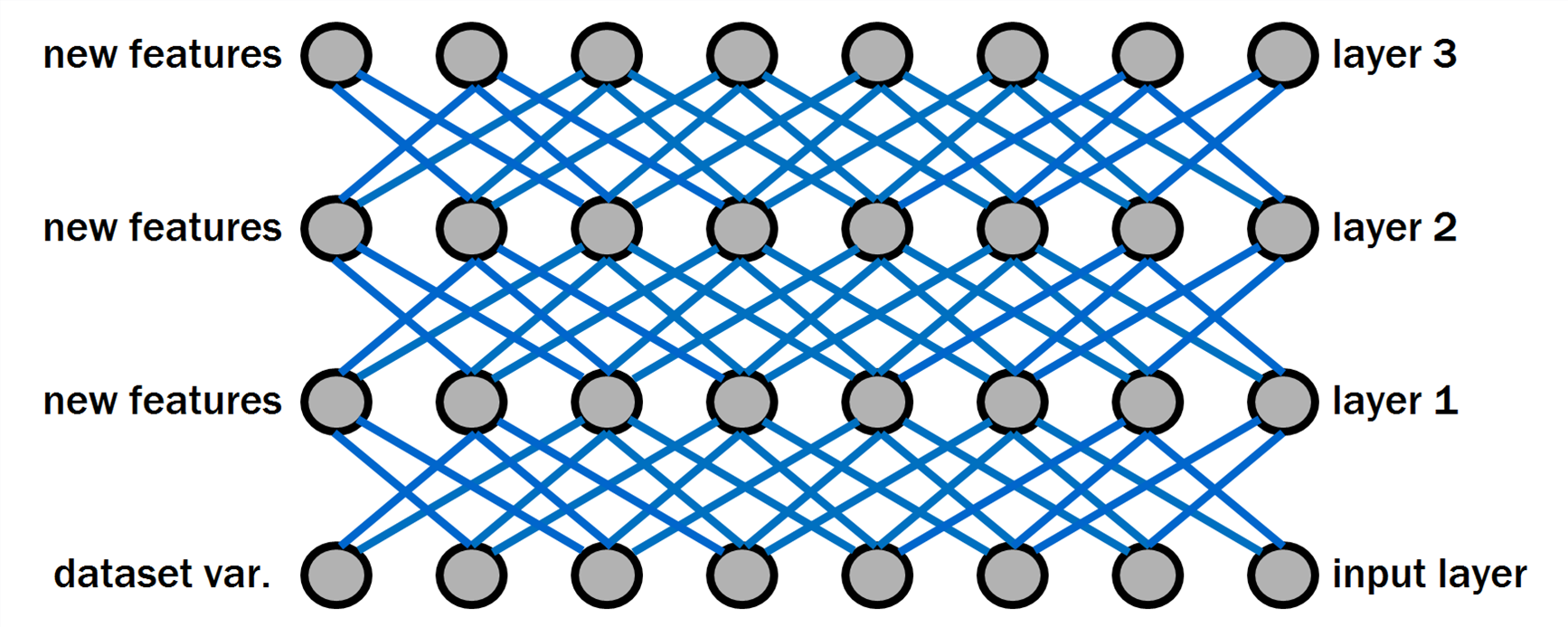}}}
\caption{New features. New variables functions of the dataset variables are created, arranged in layers. The layers are stacked one on top of the other, building a neural network.}
\label{newfeat}
\end{center} \end{figure}
 
\ifhpar \colorbox{colhd}{SDS definition} \\ \fi
The procedure to create the scrambled dataset SDS from RDS is illustrated in Figs.~\ref{rdsdef} and~\ref{sdsdef}. A generic row (example) of SDS is obtained combining one of the possible values taken by each variable on RDS, and the whole SDS is created putting together all such combinations (the frequency of occurrence of each variable's value corresponds to the frequency of occurrence observed in RDS).  

\ifhpar \colorbox{colhd}{SDS definition /2} \\ \fi
In practise, SDS is a version of RDS in which the single variables have the same probability distribution as in RDS, but where the correlations between variables are ``broken''. Since the size of SDS grows very rapidly with the number of variables and the number of possible values, in practical applications it will be necessary to use samples of SDS. The product of probabilities $P(A) \cdot P(B)$ in equation (1) can now be rewritten as $P(A \cap B)|SDS$. In this way, the numerator of equation (1) can be reformulated as:

\begin{flushleft}
$P(A \cap B)|RDS - P(A \cap B)|SDS$
\end{flushleft}

\ifhpar \colorbox{colhd}{comparison generalisation} \\ \fi
The next step to generalise equation (1) consists in replacing function $P(A \cap B)$ with an arbitrary function $F(X_{i})$ of the dataset variables $X_i$. Then, we compute the probability qr that F is TRUE on RDS and the probability qs that F is TRUE on SDS (for variables comprised in the [0,1] interval, being TRUE means being $\geq$ 0.5). Finally, we compare qr and qs. This comparison can be done with many different formulas; in our computational experiments, the following expression has proved to be effective:  

\begin{flushleft}
$hoc = 1-qs/qr = 1- \frac {P(F(X_i)=TRUE)|SDS} {P(F(X_i)=TRUE)|RDS} $ (3)
\end{flushleft}

\ifhpar \colorbox{colhd}{hoc definition} \\ \fi
We propose this quantity, called \emph{hyper-occurrence (hoc)}, as a generalised definition of correlation among multiple variables. This parameter is defined only for $qs<qr$, in which case it is comprised in the [0,1] interval. It is a measure of the difference between the distributions of values of function $F(X_{i})$ on the two datasets RDS and SDS. More sophisticated difference measures, such as the Kullback-Leibler divergence, can also be used. Other variations, such as using the mean of F instead of its TRUE-probability, are also possible.

\section{Objective function for unsupervised learning}

\begin{figure}[t]
\begin{minipage}[t]{0.47\textwidth} \centering \hspace*{-0.0cm}
\includegraphics[width=0.90\textwidth]{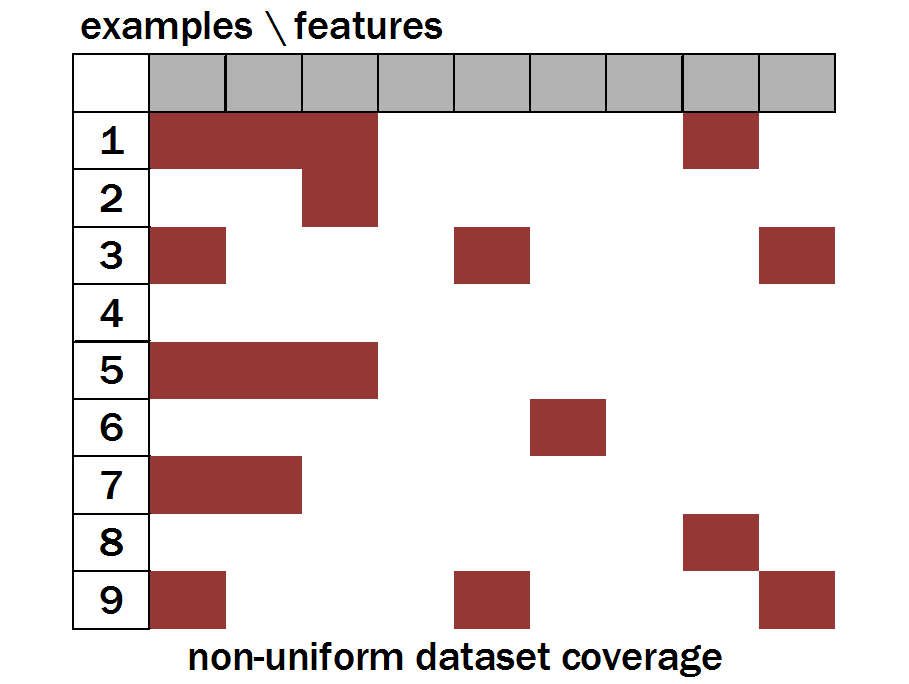}
\caption{Dataset coverage. Example of non-uniform dataset coverage (active features (value $\geq 0.5$) are marked in red, inactive ones are marked in white). Some examples are covered with many features, other examples are covered with few features or are not covered at all.}
\label{coverbad}
\end{minipage} \qquad
\begin{minipage}[t]{0.47\textwidth} \centering \hspace*{-0.0cm}
\includegraphics[width=0.90\textwidth]{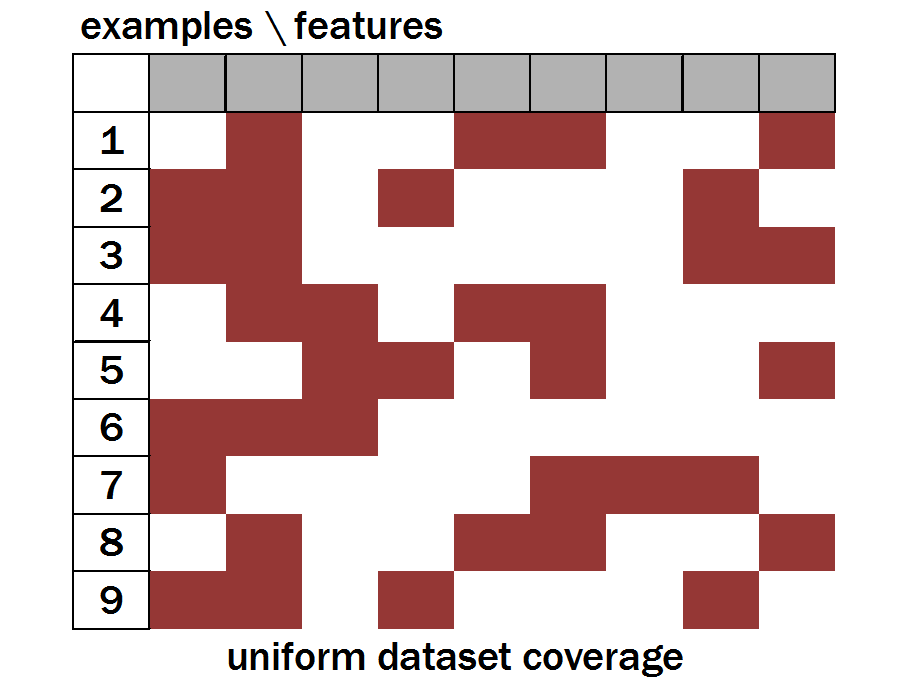}
\caption{Dataset coverage. Example of uniform dataset coverage. All examples are covered with approximately the same number of features. Features must have a high hoc value.}
\label{covergood}
\end{minipage} \qquad
\end{figure}

\ifhpar \colorbox{colhd}{neural network} \\ \fi
So far, we have figured out how to assess whether a single new variable, function of the dataset variables, captures correlation in its inputs. We can imagine to create many such variables (called features) and arrange them in layers, stack many such layers one on top of the other, and construct a neural network (Fig.~\ref{newfeat}). If we optimise the function parameters with a search algorithm based on a certain objective function, we run the risk that two or more features found by the algorithm are identical or very similar, a result we are not interested in. 

\ifhpar \colorbox{colhd}{new variables, ds coverage} \\ \fi
The second idea derives from observing that, when two features are similar, they take low and high values in correspondence of the same examples. An indirect way to foster feature diversity consists in requesting that for each example there is a high number of active features, and that this number has a low variance across the dataset. In other words, we want to avoid that for some examples there are many active features and for other examples very few (Fig.~\ref{coverbad}): the dataset ``coverage'' should be high and uniform (Fig.~\ref{covergood}).  

\ifhpar \colorbox{colhd}{procedure} \\ \fi
This result can be achieved with the following procedure. For each example e, for each feature $F_k$, we calculate the product of the feature value and the feature hoc value: $P_k(e)=F_k(e) \cdot hoc(F)$. Then, we compute the square root of the mean of the H highest $P(e)$: $d(e) = \sqrt{E_k(P_k(e))}$ (H=10 in our simulations). Thanks to the square root, low covered examples have a disproportionately high value, which tends to favour uniformity of coverage. The objective function, which we call \emph{coverage}, is the mean value of $d(e)$ on the whole RDS: $cov=E_e(d(e))$. The incorporation of hoc in the formulas guarantees that the dataset coverage is obtained through features with high hoc values.  



\section{Simulation}
\label{sec:poet}

\begin{figure}[t] \begin{center} \hspace*{-0.0cm}
{\fboxrule=0.0mm\fboxsep=0mm\fbox{\includegraphics[width=17.00cm]{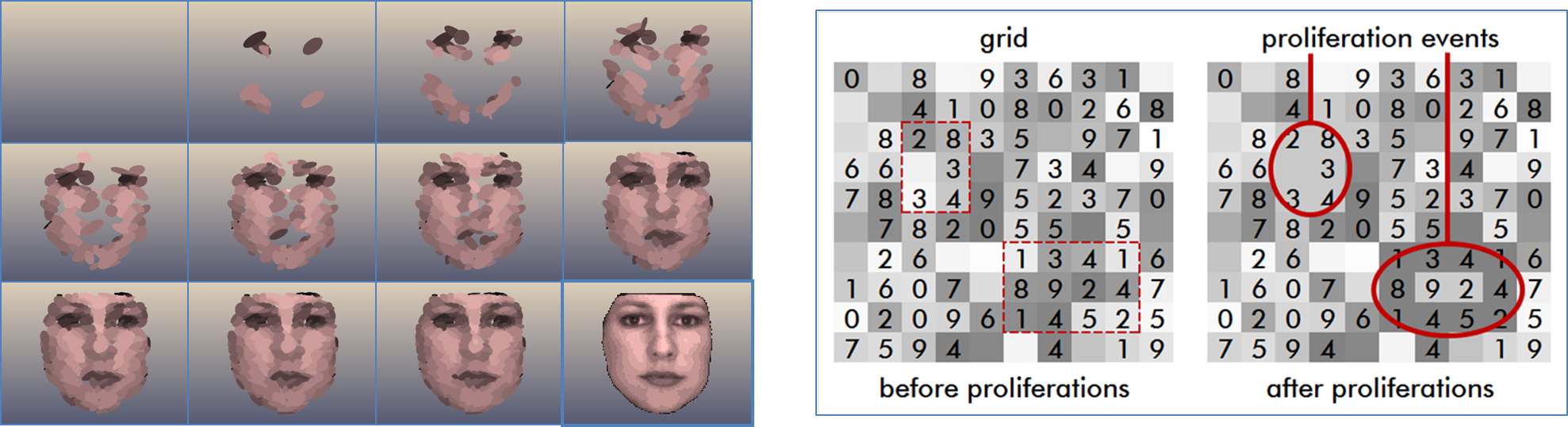}}}
\caption{On the left: example of a structure generated using ET. The sequence illustrates stages of the developmental sequence of a picture representing a face (the last frame is the target). This example suggests that ET might be a powerful algorithm to optimise large search spaces where both regularities and irregularities are present. On the right: parameter mapping in POET. To each cell two numbers are associated: a real valued number, represented by the intensity of shading, and an integer value (shown inside the cell). The value of parameter whose index value is $i$ is calculated by adding all real valued numbers of cells whose integer value is $i$.}
\label{poet}
\end{center} \end{figure}

\ifhpar \colorbox{colhd}{ET description} \\ \fi
In this work, the search of function parameters to optimise the quantities defined in the previous section will be carried out through an evolutionary algorithm called \emph{POET}, based on an evo-devo model called \emph{Epigenetic Tracking} (\emph{ET}) \citep{fontana2008}. In ET artificial bodies are composed of two categories of cells: \emph{stem cells} and \emph{normal cells}. Development starts from a set of initial cells placed on a grid and unfolds in time through \emph{developmental stages} regulated by a \emph{global clock} shared by all cells. Stem cells direct the developmental process. When a stem cell is activated, it can orchestrate either a large-scale apoptosis (death of a large number of cells in the volume around a stem cell), or a cell proliferation, filling up the volume around the original stem cell. Cells, when created, take up place in a two dimensional grid. ET can be coupled with a genetic algorithm and becomes an evo-devo process able to generate complex structures (Fig.~\ref{poet}-left).

\ifhpar \colorbox{colhd}{POET description} \\ \fi
POET (for \emph{P}arameter \emph{O}ptimization using \emph{E}pigenetic \emph{T}racking) is a search method that builds upon ET. In POET each grid point is associated to a couple of values: k and c. k represents the id number of a parameter, and c represents the parameter value. Both numbers can be modified by means of a set of change events, orchestrated by stem cells. In POET there are two types of change events: (1) proliferation, which changes the c values in the area around the activated stem cell, and (2) swap, which changes the k values. This is obtained by swapping the k values of the areas in the grid. Through a sequence of proliferation and swap events, the (c,k) values are modified in all grid positions (Fig.~\ref{poet}-right). This translates to a change of the encoded parameter set. 
 
\ifhpar \colorbox{colhd}{essentially} \\ \fi
Essentially, compared to a standard genetic algorithm, POET allows to evolve chunks of the genome which encode changes to the parameter set. As evolution progresses, older changes are ``frozen'' (they cannot be evolved anymore). In previous work \citep{fontana2014poet} POET was used to train a neural network, for a visual classification task represented by a subset of the MNIST written character dataset. The choice of an evolutionary method to search the parameter space is motivated by its generality and applicability to an arbitrary objective function.

\begin{figure}[t] \begin{center} \hspace*{-0.5cm}
{\fboxrule=0.0mm\fboxsep=0mm\fbox{\includegraphics[width=18.00cm]{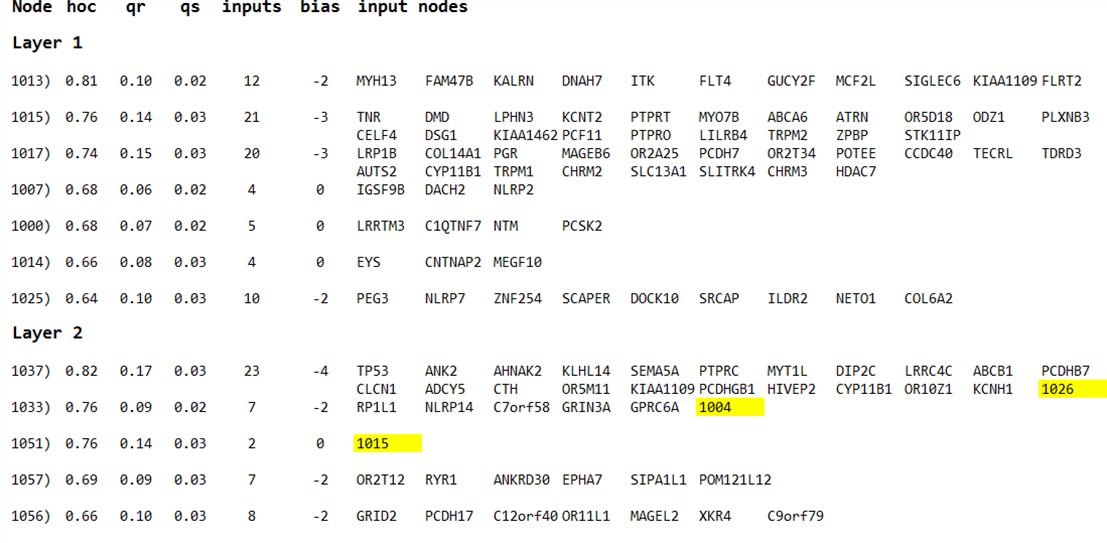}}}
\caption{Search results. The table reports the first 5 rules with the highest hoc value in layer 1 and layer 2. The field ``bias'' contains the number of genes in the subset that need to be mutated for the rule to be TRUE. Elements marked in yellow represent layer 1 nodes, which do not correspond to genes.}
\label{results}
\end{center} \end{figure}

\ifhpar \colorbox{colhd}{cancer ds} \\ \fi
The method proposed has been tested on a subset of the tumor portal dataset \citep{lawrence2013}. Each of the 100 rows of the dataset represents a patient affected by Lung Adenocarcinoma. Each of the 1000 column contains a 1 if the corresponding gene is mutated, a 0 otherwise (these 1000 genes are those with the highest mutation rate for this kind of tumour). The purpose of this test is to see if the method is able to find correlation patterns on a real-word dataset, not to gain insight into tumour biology. The rules searched have the form: if in the gene subset $\{ G_1,G_2,..., G_k\}$ at least Q genes are mutated, the rule is TRUE. This is a quite general rule that includes logical And and logical Or as special cases. 
 
\ifhpar \colorbox{colhd}{results} \\ \fi
We report the results of one simulation, in which POET has been allowed to run for 10000 generations. The network is composed of two layers (besides the input layer), each composed of 32 nodes, each node potentially fully connected to all nodes of all previous layers. Fig.~\ref{results} reports the 5 rules found with the highest hoc value for layers 1 and 2. Layer 2 rules tend to have higher qr values. The dataset coverage value is 0.17 considering nodes of layer 1 and 0.21 considering nodes of layer 1 and nodes of layer 2. These preliminary results clearly show that the method is able to discover rules with high hoc values. Further conditions can be imposed on the rules, on qr and qs values, to steer the search towards particular regions of the parameter space.

\section{Discussion}

\subsection*{Correlation}

\ifhpar \colorbox{colhd}{Pearson correlation} \\ \fi
The method described was introduced as a generalisation of the concept of Pearson correlation. The generalisation is done in two ways. First of all, while Pearson correlation involves only two variables, our method allows to detect correlation among an arbitrary number of variables. Secondly, while in Pearson's formula the probability of a specific function is compared to the product of marginal probabilities, our method allows the use of an arbitrary function. This last statement can be reformulated by saying that, while Pearson captures linear correlation, hoc captures any correlation.

\subsection*{Hypothesis testing}

\ifhpar \colorbox{colhd}{hypotest} \\ \fi
Hoc can also be seen as a generalisation of the hypothesis testing procedure. In hypothesis testing, a statistic calculated on a dataset is compared to the same statistic calculated on another dataset, which corresponds to the null hypothesis. If the difference between the two values is sufficiently large, we can reject the null hypothesis. In this context, the statistic is usually a simple function of the dataset variables; moreover, it is fixed. If the difference is too small to reject the null hypothesis, no further action is taken. In our approach, the function is variable and the function parameters are optimised the maximise the difference.     

\subsection*{Association rules}

\ifhpar \colorbox{colhd}{association rules} \\ \fi
The method described bears some resemblance to swap randomisation \citep{gionis2007}, a technique used to assess the statistical significance of itemsets found by data mining algorithms. The key difference between swap randomisation and hyper-occurrence is that in the first case the functions that capture correlation are chosen in advance and the aim of randomisation is to assess if the associations found are statistically significant. The only function used in this approach is the logical $And$ of a subset of the inputs. This reflects the intended use, which is to discover frequent itemsets in a database of transactions. 

\ifhpar \colorbox{colhd}{our approach} \\ \fi
In our approach, the difference between RDS and SDS is the driving force the shapes the choice of the functions, which can take different forms depending on a set of parameter values. Moreover, statistical significance is not the main criterion. Indeed, if the dataset is very large, even small deviations between RDS and SDS are significant: our aim is to maximise this difference regardless of significance. Another difference concerns the structure of the randomised dataset, which in \citep{gionis2007} has the same row and column margins as the original dataset, while in our case only the column margin is maintained. Finally, in our approach the layers of functions are stacked one on top of the other, building a neural network. We like to think of this work as a joint linking together the fields of statistics, data mining and neural networks. 

\subsection*{Features for neural networks}

\ifhpar \colorbox{colhd}{deep learning, convolutional} \\ \fi
The state of the art in the field of neural networks is represented by convolutional networks \citep{Krizhevsky2012} trained with back-propagation, which appears to be immune from the vanishing gradient problem \citep{hochreiter2004grad} when used with huge amounts of data. A recent method achieves an error rate of 3.57\% on the ImageNet dataset using a network with 152 layers \citep{kaiming2015}. However, in spite of the successes recorded, several issues remain unresolved.

\ifhpar \colorbox{colhd}{dl issues, too many examples required} \\ \fi
A first problem is that neural networks require a very large number of training examples, while human beings, on their hand, seem to be able to form new concepts with a more limited exposure to data. Another issue is overfitting, which occurs when a complex model (many parameters) is used to interpret a small dataset (few examples). The model ends up describing random error or noise instead of the relationship among variables, which translates to poor performance on unseen data. Several methods, such as regularisation \citep{girosi1995regul}, have been proposed to reduce overfitting, and new ones keep being proposed \citep{srivastava2014dropout}.

\ifhpar \colorbox{colhd}{dl issues, adversarial examples} \\ \fi
Adversarial examples are examples of images which appear as random noise to human observers, but that neural networks label with high confidence as belonging to one of the classes used for training \citep{nguyen2015adex}. Adversarial examples can be constructed by tweaking some pixels in a random image. This slightly depressing phenomenon has been explained by the presence of the long ``shadow'' cast by the network in regions of the weight space unexplored during training, in which the network cannot be used to make reliable predictions.

\ifhpar \colorbox{colhd}{improvements for NN} \\ \fi
The method proposed discovers features that are statistically robust: once this robustness is validated on the small subset of the dataset, it is expected to extend to the entire dataset. This could enable learning with fewer examples and provide protection against overfitting. Finally, since the features discovered have low occurrence on SDS, a classificator based on such features holds the promise of being immune from random-looking examples, such as the recently discovered adversarial examples.

\section{Conclusions}

\ifhpar \colorbox{colhd}{xxxxxxxx} \\ \fi
In this work we have described a new method to discover correlation patterns among an arbitrary number of variables. The method is based on a node-based statistical criterion called hyper-occurrence, and on a layer-based criterion which fosters node diversity through dataset coverage. The parameter search is carried out with a special genetic algorithm called POET, it is however susceptible to be done with any suitable search algorithm. Future work will be aimed to further investigate the theoretical basis of the method, as well as to explore applications to other real world datasets.

\section{Disclaimer}

The author is an employee of the European Research Council Executive Agency. The views expressed are purely those of the writer and may not in any circumstances be regarded as stating an official position of the European Commission.

\bibliographystyle{apalike}
\bibliography{lsensfiltr}
 
\end{document}